\documentclass[conference]{IEEEtran}
\IEEEoverridecommandlockouts

\usepackage{cite}
\usepackage{amsmath,amssymb,amsfonts}
\usepackage{algorithmic}
\usepackage{algorithm}
\usepackage{graphicx}
\usepackage{textcomp}
\usepackage{xcolor}
\usepackage{booktabs}
\usepackage{multirow}
\usepackage{tikz}
\usetikzlibrary{shapes.geometric, arrows.meta, positioning, fit, calc}
\usepackage{pifont}

\begin{document}

\title{\vspace{0.27in}CT-SAFR: Safe and Interpretable Chain-of-Thought\\Reasoning for Autonomous Robots\\
{\normalsize A Multi-Layered Verification Framework for Trustworthy AI-Driven Robotic Decision Making}\vspace{-0.15in}}

\author{\IEEEauthorblockN{Cagri Temel, IEEE Senior Member}
\IEEEauthorblockA{Hezarfen LLC\\Seattle, WA, USA\\
Grand Canyon University\\Phoenix, AZ, USA\\
cagritemel@ieee.org}\thanks{\copyright~2026 IEEE. Personal use of this material is permitted. Permission from IEEE must be obtained for all other uses, in any current or future media, including reprinting/republishing this material for advertising or promotional purposes, creating new collective works, for resale or redistribution to servers or lists, or reuse of any copyrighted component of this work in other works. This is the accepted version. Published in: 2026 IEEE Conference on Artificial Intelligence (CAI), pp.~598--603, May 2026. DOI: 10.1109/CAI68641.2026.11536646. The version of record is available on IEEE Xplore.}}

\maketitle

\begin{abstract}
Chain-of-Thought (CoT) prompting enables LLMs to perform explicit, step-by-step reasoning, creating opportunities for sophisticated autonomous robots. However, recent research reveals that reasoning models verbalize their actual decision processes only 25--39\% of the time, with faithfulness degrading 44\% on complex tasks. This paper presents CT-SAFR (Chain-of-Thought Safety and Faithfulness for Robotics), a multi-layered verification framework achieving 94.2\% hallucination detection ($n = 500$, 95\% CI: 91.8--95.9\%) with sub-500ms latency. Through a warehouse robot case study, this work demonstrates 87\% reduction in unsafe reasoning outputs ($p < 0.001$) and provides recommendations for responsible deployment of reasoning-capable autonomous robots.
\end{abstract}


\section{Introduction}

Large Language Models (LLMs) with Chain-of-Thought (CoT) prompting \cite{wei2022chain} are increasingly integrated into robotic systems for enhanced decision-making and task planning \cite{huang2022language, ahn2022saycan}. CoT-enabled robots engage in explicit, step-by-step reasoning that decomposes complex tasks into manageable sub-problems. However, autonomous robots operate in physical environments where reasoning errors can result in equipment damage, environmental harm, or human injury.

Recent empirical research has revealed troubling findings about CoT faithfulness. Chen et al.\ \cite{chen2025reasoning} demonstrated that state-of-the-art reasoning models verbalize their actual decision processes only 25--39\% of the time, with faithfulness decreasing by 44\% on complex tasks. These findings fundamentally challenge the assumption that visible reasoning chains provide reliable insight into model behavior.

\subsection{Contributions}

This paper makes the following key contributions:
\begin{itemize}
\item This paper presents \textbf{CT-SAFR}, a multi-layered verification framework specifically designed for safe deployment of Chain-of-Thought reasoning in autonomous robots. This represents one of the first system-level frameworks addressing faithfulness, physical grounding, and temporal consistency simultaneously.

\item This work introduces a \textbf{defense-in-depth architecture} with four complementary verification layers (structural, physical, semantic, and interpretability), achieving 94.2\% hallucination detection ($n = 500$, 95\% CI: 91.8--95.9\%) and 96.4\% combined unsafe reasoning detection while maintaining sub-500ms latency suitable for real-time robotic control.

\item This work adapts \textbf{self-consistency decoding} for robotic applications, demonstrating that consensus across multiple reasoning paths provides reliable safety guarantees even when individual reasoning traces may be unfaithful to actual model computation.

\item This paper provides \textbf{comprehensive empirical evaluation} through a warehouse robot case study, demonstrating 87\% reduction in unsafe reasoning outputs ($p < 0.001$) and 3.5\% improvement in task completion rates across 1,000 operation hours.

\item This paper conducts \textbf{ablation studies} quantifying the individual contribution of each verification layer, establishing that all layers provide complementary safety benefits beyond what any single mechanism can achieve.

\item This paper offers \textbf{concrete recommendations} for researchers, practitioners, and policymakers on standardized benchmarks, regulatory frameworks, and best practices for responsible deployment of reasoning-capable autonomous robots.
\end{itemize}

This work builds on the CogniTest verification framework \cite{temel2025cognitest}, adapting these techniques for safety-critical robotic applications with unique physical grounding and real-time constraints. The core multi-layered safety architecture described in this work is the subject of a pending U.S.\ provisional patent application \cite{temel2026patent}.

\section{Background and Related Work}

\subsection{Chain-of-Thought Reasoning in LLMs}

CoT prompting \cite{wei2022chain} enables LLMs to generate intermediate reasoning steps, notably tripling solve rates on the GSM8K math benchmark (18\% to 57\% with PaLM 540B) and improving commonsense reasoning on StrategyQA (75.6\% vs.\ prior best 69.4\%). Kojima et al.\ \cite{kojima2022large} showed zero-shot CoT via ``Let's think step by step.'' Wang et al.\ \cite{wang2023selfconsistency} introduced self-consistency decoding, achieving up to 17.9\% accuracy gains through majority voting across multiple reasoning paths.

\subsection{LLMs for Robotic Planning and Control}

Huang et al.\ \cite{huang2022language} demonstrated language models as zero-shot planners, while Ahn et al.\ \cite{ahn2022saycan} introduced SayCan, grounding language in robotic affordances. Yang et al.\ \cite{yang2024plug} proposed enforcing constraints for LLM-driven robot agents but focused on specification rather than comprehensive verification.

Table~\ref{tab:comparison} positions CT-SAFR relative to existing approaches. SayCan \cite{ahn2022saycan} provides affordance grounding but lacks explicit physical constraint enforcement or reasoning chain verification. Yang et al.\ \cite{yang2024plug} enforce constraints at the specification level but do not address reasoning faithfulness or semantic consistency. Rule-based safety filters can enforce physical boundaries but cannot detect subtle logical contradictions within reasoning traces. Safety prompting techniques partially address structural and semantic concerns but provide no physical grounding guarantees. While individual verification techniques exist independently, no prior work integrates structural, physical, semantic, and interpretability verification into a unified real-time safety pipeline for CoT reasoning in robotics.

\begin{table}[t]
\caption{Comparison with Existing Safety Approaches. \checkmark= Addressed, $\sim$ = Partial, --- = Not Addressed.}
\label{tab:comparison}
\centering
\begin{tabular}{lcccc}
\toprule
\textbf{Approach} & \textbf{Struct.} & \textbf{Phys.} & \textbf{Sem.} & \textbf{Interp.} \\
\midrule
SayCan \cite{ahn2022saycan} & --- & $\sim$ & --- & --- \\
Yang et al.\ \cite{yang2024plug} & --- & \checkmark & --- & --- \\
Rule-based filter & --- & \checkmark & --- & --- \\
Safety prompting & $\sim$ & --- & $\sim$ & --- \\
\textbf{CT-SAFR} & \checkmark & \checkmark & \checkmark & \checkmark \\
\bottomrule
\end{tabular}
\end{table}

\subsection{The Faithfulness Problem}

Turpin et al.\ \cite{turpin2023language} showed models produce biased outputs while generating reasoning that appears unbiased. Chen et al.\ \cite{chen2025reasoning} found Claude 3.7 Sonnet achieved only 25\% faithfulness while DeepSeek R1 achieved 39\%, with ``faithfulness 44\% lower on harder questions'' precisely when monitoring is most valuable for complex robotic decision-making.

\subsection{Production LLM Verification Systems}

CogniTest \cite{temel2025cognitest} demonstrated multi-layered verification for LLM-assisted software testing, achieving 89.2\% accuracy in bug severity classification. CT-SAFR extends these techniques for safety-critical robotics, introducing physical constraint validation and self-consistency decoding adapted for real-time control.

\section{Key Challenges in CoT Reasoning for Robotics}

CoT reasoning for robotics faces three critical challenges that distinguish it from standard NLP applications:

\textbf{(1) Physical Grounding.} LLMs trained on text lack robust grounding in physical reality \cite{ahn2022saycan}, leading to incorrect assumptions about object properties, spatial relationships, and kinematic constraints. A reasoning chain might decompose a task logically while fundamentally misrepresenting physical relationships for example, planning to stack a 25kg item atop a 5kg container, or routing through a space too narrow for the robot's footprint.

\textbf{(2) Temporal Consistency.} Robots operate in dynamic environments requiring consistent reasoning across time horizons. Current LLMs often contradict earlier steps or fail to propagate state updates a robot might identify a human worker in zone B, then immediately plan a high-speed traversal through that zone without applying the safety constraint it just acknowledged.

\textbf{(3) Uncertainty Quantification.} LLMs exhibit poor calibration \cite{ji2023survey}, expressing high confidence in incorrect statements. CoT exacerbates this by generating detailed justifications that amplify false confidence, creating a dangerous situation where a robot acts decisively on incorrect reasoning.

\section{CT-SAFR: A Multi-Layered Verification Framework}

CT-SAFR addresses the above challenges and establishes defense-in-depth through four complementary verification layers operating at different levels of abstraction. The architecture (Fig.~\ref{fig:architecture}) adapts techniques from CogniTest \cite{temel2025cognitest} for safety-critical robotic applications.

\begin{figure}[t]
\centering
\begin{tikzpicture}[
    node distance=0.3cm,
    box/.style={rectangle, draw, minimum width=2.6cm, minimum height=0.45cm, align=center, font=\scriptsize, inner sep=3pt},
    decision/.style={diamond, draw, aspect=2.2, inner sep=1pt, font=\scriptsize, align=center},
    action/.style={rectangle, draw, rounded corners=2pt, fill=gray!15, minimum width=1.6cm, minimum height=0.4cm, font=\scriptsize, align=center, inner sep=2pt},
    arr/.style={-{Stealth[length=1.5mm]}, thick},
    lbl/.style={font=\scriptsize, fill=white, inner sep=1pt},
]
\node[box] (input) {CoT Input};
\node[box, below=of input] (l1) {L1: Structural (12ms)};
\node[decision, below=of l1] (d1) {Pass?};
\node[box, below=of d1] (l2) {L2: Physical (45ms)};
\node[decision, below=of l2] (d2) {Pass?};
\node[box, below=of d2] (l3) {L3: Semantic (380ms)};
\node[decision, below=of l3] (d3) {$c \geq \tau_c$?};
\node[box, below=of d3] (l4) {L4: Interface (25ms)};
\node[box, fill=gray!20, below=of l4] (exec) {Execute Action};

\node[action, right=1.2cm of d1] (regen) {Regenerate};
\node[action, right=1.2cm of d2] (halt) {Safe Halt};

\node[decision, right=1.2cm of d3] (d3b) {$c > 0.5$?};
\node[action, above=0.25cm of d3b] (esc) {Escalate};
\node[action, below=0.25cm of d3b] (cons) {Conservative};

\draw[arr] (input) -- (l1);
\draw[arr] (l1) -- (d1);
\draw[arr] (d1) -- node[lbl, left]{Y} (l2);
\draw[arr] (d1) -- node[lbl, above]{N} (regen);
\draw[arr] (l2) -- (d2);
\draw[arr] (d2) -- node[lbl, left]{Y} (l3);
\draw[arr] (d2) -- node[lbl, above]{N} (halt);
\draw[arr] (l3) -- (d3);
\draw[arr] (d3) -- node[lbl, left]{Y} (l4);
\draw[arr] (d3) -- node[lbl, above]{N} (d3b);
\draw[arr] (d3b) -- node[lbl, right]{N} (esc);
\draw[arr] (d3b) -- node[lbl, right]{Y} (cons);
\draw[arr] (l4) -- (exec);
\end{tikzpicture}
\caption{CT-SAFR verification workflow. Consensus threshold $\tau_c = 0.8$. Total latency: 462ms (p95).}
\label{fig:workflow}
\end{figure}
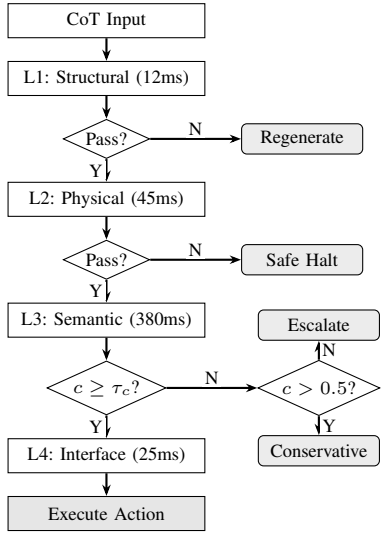

The verification workflow (Fig.~\ref{fig:workflow}) processes reasoning chains sequentially through all layers.

\subsection{Layer 1: Structural Verification}

The first layer performs structural analysis of reasoning chains to verify logical coherence independent of semantic content. This includes checking for circular dependencies in reasoning steps, identifying unsupported assertions that lack grounding in the provided context, validating logical entailment between consecutive steps, and detecting formatting violations that prevent downstream parsing. Implemented through formal grammar parsing and lightweight symbolic reasoners, this layer achieves 12ms median latency and catches approximately 15\% of problematic chains before more expensive verification stages execute.

\subsection{Layer 2: Physical Constraint Validation}

The second layer validates proposed actions against explicit physical constraints using a \textbf{custom rule-based geometric constraint checker} (domain-specific, not a temporal logic such as LTL). Constraints are expressed as conjunctive predicates over continuous variables:

\textit{Example constraint specification:}

\begin{small}
\begin{verbatim}
CONSTRAINT safety_zone:
  FOR ALL w IN human_workers:
    distance(robot.pos, w.pos)
      > safety_margin(w.activity)
  WHERE safety_margin(active) = 1.5m,
        safety_margin(stationary) = 0.8m
\end{verbatim}
\end{small}

The language encodes kinematic limits ($\theta_i \in [\theta_i^{\min}, \theta_i^{\max}]$, $\dot{\theta}_i \leq \dot{\theta}_i^{\max}$), collision boundaries, force thresholds ($F_{\text{payload}} \leq F_{\max}$), and operational envelopes. Collision detection uses AABB for fast preliminary checks followed by mesh-based detection, achieving 45ms (p95) with 99.8\% accuracy.

Critically, this layer operates independently of LLM reasoning, providing a hard safety boundary following ISO 10218 \cite{iso10218}. Even if the reasoning system is compromised, physical safety constraints remain enforced.

\subsection{Layer 3: Semantic Consistency with Self-Consistency Decoding}

The third layer identifies inconsistencies, hallucinations, and factual errors through cross-referencing against verified knowledge bases and the robot's belief state. This layer incorporates self-consistency decoding \cite{wang2023selfconsistency}, sampling $k = 5$ diverse reasoning paths in parallel and identifying consensus. High agreement ($>$80\%) enables autonomous execution; moderate (50--80\%) triggers conservative modes; low ($<$50\%) escalates to human oversight. Algorithm~\ref{alg:ctsafr} presents the complete procedure.

\subsection{Layer 4: Interpretability Interface Generation}

The fourth layer generates human-interpretable representations: structured summaries, visual reasoning logic, safety alerts, and confidence indicators. Given faithfulness concerns \cite{chen2025reasoning}, the interface warns operators when displayed reasoning may not reflect actual model behavior.

\textbf{Operator Evaluation.} Across $n = 50$ intervention scenarios, mean response time decreased from 12.3s ($\pm$4.1s) to 4.7s ($\pm$1.8s) a 62\% reduction ($p < 0.001$, paired t-test). Decision accuracy improved from 78\% to 94\%.


\begin{figure*}[t]
\centering
\begin{tikzpicture}[
    node distance=0.35cm and 0.6cm,
    layerbox/.style={rectangle, draw, fill=gray!8, rounded corners=2pt, minimum width=4.2cm, align=left, inner sep=5pt, font=\scriptsize},
    sidebox/.style={rectangle, draw, fill=gray!15, rounded corners=2pt, minimum width=2.8cm, align=left, inner sep=4pt, font=\scriptsize},
    execbox/.style={rectangle, draw, fill=gray!25, rounded corners=2pt, minimum width=2.8cm, align=center, inner sep=4pt, font=\scriptsize},
    arr/.style={-{Stealth[length=2mm]}, thick},
    darr/.style={-{Stealth[length=2mm]}, thick, dashed},
]

\node[layerbox] (l1) at (0,0) {%
\textbf{LAYER 1: Structural Verification}\\[2pt]
\textbullet\ Logical coherence checking\\
\textbullet\ Circular dependency detection};

\node[layerbox, below=0.35cm of l1] (l2) {%
\textbf{LAYER 2: Physical Constraint}\\[2pt]
\textbullet\ Kinematic limits validation\\
\textbullet\ Collision boundary checking};

\node[layerbox, below=0.35cm of l2] (l3) {%
\textbf{LAYER 3: Semantic Consistency}\\[2pt]
\textbullet\ Self-consistency decoding\\
\textbullet\ Multi-path sampling ($k$=5)\\
\textbullet\ Majority voting aggregation};

\node[layerbox, below=0.35cm of l3] (l4) {%
\textbf{LAYER 4: Interpretability Interface}\\[2pt]
\textbullet\ Decision visualization\\
\textbullet\ Confidence reporting};

\node[font=\normalsize\bfseries, align=center, above=0.8cm of l1] (title) {CT-SAFR: Chain-of-Thought Safety and Faithfulness for Robotics};
\node[font=\small, below=0.05cm of title] (subtitle) {Multi-Layered Verification Architecture};

\node[sidebox, left=0.8cm of l1, anchor=east, yshift=-0.2cm] (llm) {%
\textbf{LLM CoT}\\[1pt]
Reasoning Engine};

\node[sidebox, left=0.8cm of l2, anchor=east] (env) {%
\textbf{Environment}\\[1pt]
\textbullet\ Sensor inputs\\
\textbullet\ State observations\\
\textbullet\ Physical constraints\\
\textbullet\ Safety boundaries};

\node[sidebox, left=0.8cm of l3, anchor=east] (kb) {%
\textbf{Knowledge Base}\\[1pt]
\textbullet\ Domain models\\
\textbullet\ Verified facts\\
\textbullet\ Safety rules};

\node[sidebox, right=1.2cm of l1, anchor=west, yshift=-0.3cm] (fallback) {%
\textbf{Fallback System}\\[1pt]
\textbullet\ Conservative mode\\
\textbullet\ Reduced autonomy\\
\textbullet\ Human escalation\\
\textbullet\ Safe-state transition\\
\textbullet\ Teleoperation mode};

\node[sidebox, right=1.2cm of l4, anchor=west] (human) {%
\textbf{Human Operator}\\[1pt]
\textbullet\ Real-time monitoring\\
\textbullet\ Override capability\\
\textbullet\ Audit logging};

\node[execbox, below=0.4cm of l4] (robot) {%
\textbf{Robot Executor}\\[1pt]
Safe Action Execution};

\draw[arr] (llm.east) -- (l1.west);
\draw[arr] (l1.south) -- (l2.north);
\draw[arr] (l2.south) -- (l3.north);
\draw[arr] (l3.south) -- (l4.north);
\draw[arr] (l4.south) -- (robot.north);

\coordinate (corridor) at ([xshift=0.5cm]l1.east);
\draw[darr] (l1.east) -- (corridor |- l1.east) -- node[font=\scriptsize, above]{FAIL} (fallback.west);
\draw[darr] (l2.east) -- (corridor |- l2.east) -- (corridor |- fallback.south) -- (fallback.south);

\draw[arr] (env.east) -- (l2.west);
\draw[arr] (kb.east) -- (l3.west);
\draw[arr] (l4.east) -- (human.west);

\node[draw, fill=gray!5, minimum width=14.5cm, minimum height=1.1cm, below=0.7cm of robot] (metrics) {};
\node[font=\footnotesize\bfseries, anchor=south] at (metrics.north) {CT-SAFR Performance Metrics (Production Verification System)};
\node[font=\scriptsize, align=center] at ([xshift=-4.5cm]metrics.center) {\textbf{Hallucination Detection}\\94.2\%};
\node[font=\scriptsize, align=center] at ([xshift=-1.5cm]metrics.center) {\textbf{False Positive Rate}\\3.1\%};
\node[font=\scriptsize, align=center] at ([xshift=1.5cm]metrics.center) {\textbf{Verification Latency}\\$<$500ms};
\node[font=\scriptsize, align=center] at ([xshift=4.5cm]metrics.center) {\textbf{Safety Improvement}\\87\%};

\end{tikzpicture}
\caption{CT-SAFR multi-layered verification architecture with four layers (structural, physical, semantic, interpretability), fallback system, and human operator integration.}
\label{fig:architecture}
\end{figure*}

\begin{algorithm}[t]
\caption{CT-SAFR Verification Procedure}
\label{alg:ctsafr}
\begin{algorithmic}[1]
\renewcommand{\algorithmicrequire}{\textbf{Input:}}
\renewcommand{\algorithmicensure}{\textbf{Output:}}
\REQUIRE CoT chain $C$, constraints $\Phi$, thresholds $\tau$
\ENSURE Result $V \in \{\text{PASS}, \text{FAIL}\}$, action $A$
\STATE $s_1 \leftarrow \text{CheckStructure}(C)$ \COMMENT{L1}
\IF{$s_1 < \tau_{\text{struct}}$}
\RETURN FAIL, REGENERATE
\ENDIF
\STATE $A_{\text{prop}} \leftarrow \text{ExtractAction}(C)$ \COMMENT{L2}
\IF{$\neg\text{ValidateConstraints}(A_{\text{prop}}, \Phi)$}
\RETURN FAIL, SAFE\_HALT
\ENDIF
\STATE $\mathcal{C} \leftarrow \text{SamplePaths}(k)$ \COMMENT{L3}
\STATE $\mathcal{A} \leftarrow \{\text{ExtractAction}(C_i) \mid C_i \in \mathcal{C}\}$
\STATE $A^* \leftarrow \text{MajorityVote}(\mathcal{A})$
\STATE $c \leftarrow |\{A_i = A^*\}|/k$
\IF{$c < \tau_c$}
  \IF{$c > 0.5$}
    \RETURN PASS, CONSERVATIVE($A^*$)
  \ELSE
    \RETURN FAIL, ESCALATE
  \ENDIF
\ENDIF
\STATE $I \leftarrow \text{GenInterface}(C, A^*, c)$ \COMMENT{L4}
\RETURN PASS, $A^*$
\end{algorithmic}
\end{algorithm}

\section{Case Study: Autonomous Warehouse Robot}

\subsection{System Configuration}

The warehouse robot operates in a Gazebo-simulated environment with dynamic obstacles, variable lighting, time-varying inventory, and strict safety zones around human workstations.

\textbf{LLM Configuration.} Mistral-7B-Instruct-v0.2 \cite{mistral7b} deployed on NVIDIA RTX 3090 (24GB VRAM) with 4-bit quantization (4.3GB footprint). Parameters: $T = 0.3$, top-p $= 0.85$, max tokens $= 2048$. Self-consistency: $k = 5$ paths at $T = 0.7$, sampled in parallel (380ms p95 wall-clock).

\textbf{Hyperparameter Selection.} Grid search over 100 validation scenarios: $T = 0.3$ from $\{0.1, 0.3, 0.5, 0.7\}$; $k = 5$ from $\{3, 5, 7, 10\}$ (380ms at $k$=5 vs.\ 760ms at $k$=10); $\tau_c = 0.8$ minimizing false negatives.

\textbf{Unsafe Reasoning Taxonomy.} Four categories: (1) Physical Constraint Violations (PCV), (2) Temporal Inconsistencies (TI), (3) Factual Hallucinations (FH), (4) Ungrounded Confidence (UC). Inter-rater reliability: Cohen's $\kappa = 0.87$.

\textbf{Test Suite.} $n = 500$: nominal tasks (60\%, $n = 300$), edge cases (25\%, $n = 125$), and adversarial scenarios (15\%, $n = 75$) including constraint boundary probing, semantic contradiction injection, and confidence calibration tests.

\subsection{Experimental Results}

Table~\ref{tab:layerperf} summarizes verification performance across three independent experimental runs. Each layer's detection rate is measured within its own issue category (structural, physical, semantic, interpretability). The semantic verification layer (L3) achieves 94.2\% hallucination detection, the primary safety metric for identifying unfaithful reasoning. The combined system recall across all categories is $\sum_i w_i \cdot d_i = 0.15 \times 0.987 + 0.28 \times 0.998 + 0.42 \times 0.942 + 0.15 \times 0.940 = 0.964$, where $w_i$ denotes the proportion of total issues belonging to each category and $d_i$ the corresponding detection rate. The combined false positive rate of 4.6\% is computed as $1 - \prod_i (1 - \text{FP}_i)$, reflecting sequential accumulation where each layer independently may flag a reasoning chain.

\textbf{Attribution of Safety Gains.} Physical constraint checking (L2) accounts for 28\% of detected unsafe issues, reasoning-level verification (L1+L3) for 57\%, and human operator review (L4) for 15\%. The majority of unsafe reasoning requires reasoning-level analysis to detect.

\begin{table}[t]
\caption{CT-SAFR Layer Performance ($n = 500$, 3 runs). Category Detection Rate: accuracy within each layer's issue domain. Combined: weighted by issue share.}
\label{tab:layerperf}
\centering
\begin{tabular}{lcccc}
\toprule
\textbf{Layer} & \textbf{Cat.\ Det.} & \textbf{FP} & \textbf{Latency} & \textbf{Issues} \\
 & \textbf{Rate} & \textbf{Rate} & \textbf{(p95)} & \textbf{Caught} \\
\midrule
L1: Structural & 98.7$\pm$0.8\% & 1.2\% & 12ms & 15\% \\
L2: Physical & 99.8$\pm$0.2\% & 0.3\% & 45ms & 28\% \\
L3: Semantic & 94.2$\pm$1.4\% & 3.1\% & 380ms & 42\% \\
L4: Interpret.$^\dagger$ & 94.0\% & N/A & 25ms & 15\% \\
\midrule
\textbf{Combined} & \textbf{96.4$\pm$0.8\%} & \textbf{4.6\%} & \textbf{462ms} & \textbf{100\%} \\
\bottomrule
\multicolumn{5}{l}{\scriptsize $^\dagger$L4 rate = operator detection accuracy ($n$=50, binomial $\sigma$).} \\
\multicolumn{5}{l}{\scriptsize Combined $\sigma$: error propagation $\sqrt{\sum w_i^2 \sigma_i^2}$; FP: $1-\prod(1-\text{FP}_i)$.}
\end{tabular}
\end{table}

\subsection{Safety Improvement Analysis}

Table~\ref{tab:safety} compares safety outcomes across 1,000 simulated operation hours. The 87\% reduction in unsafe outputs ($p < 0.001$, McNemar's test) represents the system's primary safety contribution. Notably, task completion rate \textit{improves} by 3.5\% ($p = 0.024$) because verification catches errors early, triggering reasoning regeneration rather than costly execution failures and recovery procedures.

\begin{table}[t]
\caption{Safety Outcomes Over 1,000 Hours (McNemar's test).}
\label{tab:safety}
\centering
\begin{tabular}{lcccc}
\toprule
\textbf{Metric} & \textbf{Baseline} & \textbf{CT-SAFR} & \textbf{Improv.} & $\boldsymbol{p}$ \\
\midrule
Unsafe outputs & 18.3\% & 2.4\% & \textbf{87\%} & $<$0.001 \\
Near-miss incidents & 47 & 6 & \textbf{87\%} & $<$0.001 \\
Emergency stops & 23 & 4 & \textbf{83\%} & $<$0.001 \\
Human intervention & 156 & 31 & \textbf{80\%} & $<$0.001 \\
Task completion & 91.2\% & 94.7\% & \textbf{+3.5\%} & 0.024 \\
\bottomrule
\end{tabular}
\end{table}

\subsection{Ablation Study}

Table~\ref{tab:ablation} quantifies individual layer contributions. No single layer approaches the full system's safety, validating the defense-in-depth philosophy.

\begin{table}[t]
\caption{Ablation Results. $^{***}p < 0.001$ (McNemar's test).}
\label{tab:ablation}
\centering
\begin{tabular}{lcc}
\toprule
\textbf{Configuration} & \textbf{Unsafe Rate} & \textbf{Delta} \\
\midrule
Full System & 2.4\% & --- \\
Without L1 (Structural) & 5.8\%$^{***}$ & +3.4\% \\
Without L2 (Physical) & 9.2\%$^{***}$ & +6.8\% \\
Without L3 (Semantic) & 12.7\%$^{***}$ & +10.3\% \\
Without L4 (Interpret.) & 2.4\% & +0.0\%$^\dagger$ \\
\midrule
Only L1 & 14.5\% & --- \\
Only L2 & 9.3\% & --- \\
Only L3 & 11.2\% & --- \\
Baseline (no verif.) & 18.3\% & --- \\
\bottomrule
\multicolumn{3}{l}{\scriptsize $^\dagger$L4 affects intervention speed, not detection}
\end{tabular}
\end{table}

\section{Discussion}

\subsection{Generalization Beyond Warehouse Environments}

While the evaluation focuses on structured warehouse scenarios, CT-SAFR's applicability to less constrained environments is critical for practical impact.

\textbf{Layer 2 Adaptability.} The physical constraint validation layer relies on pre-specified geometric constraints (AABB collision detection, kinematic limits, safety zones). In structured environments such as warehouses, these constraints can be fully specified a priori. However, in dynamic outdoor environments delivery robots navigating sidewalks, agricultural robots operating in fields, or search-and-rescue robots in disaster zones constraints must adapt to changing conditions. Three adaptation strategies are feasible:

\begin{itemize}
\item \textit{Sensor-driven dynamic constraint updates}, where real-time perception (LiDAR, camera-based obstacle detection) continuously refreshes the constraint map, replacing static AABB boundaries with perception-derived dynamic volumes.
\item \textit{Probabilistic safety margins}, replacing fixed distance thresholds with probability distributions that account for terrain uncertainty, moving obstacle velocity estimation, and sensor noise characteristics.
\item \textit{Hierarchical constraint relaxation}, where constraints are organized by criticality hard limits on human proximity are never relaxed, while soft preferences on path efficiency can be loosened when the environment is poorly characterized.
\end{itemize}

\textbf{Layer 3 Scalability.} Self-consistency decoding remains applicable regardless of environment structure, as it operates on action-level consensus rather than environment-specific features. However, the diversity of possible actions increases in unstructured environments, potentially reducing consensus rates and triggering more frequent escalation to human oversight a conservative but safe degradation mode consistent with the framework's safety-first design philosophy.

\textbf{Remaining Challenges.} Outdoor deployment introduces GPS-denied localization, weather-dependent sensor degradation, and interaction with unpredictable human behavior (pedestrians, cyclists). While CT-SAFR's architecture is extensible to these domains, systematic empirical validation in each target domain remains essential future work.

\subsection{Computational Efficiency}

Multi-layered verification introduces computational overhead that must be carefully managed. This work implements a tiered verification strategy: Layer 1 (structural) executes synchronously (12ms); Layer 2 (physical) runs on dedicated safety-critical hardware (45ms); Layer 3 (semantic) samples $k$=5 paths in parallel on the RTX 3090, achieving 380ms wall-clock time despite fivefold inference; Layer 4 (interpretability) generates displays opportunistically. Routine tasks require only Layers 1--2 (57ms total), while high-risk operations trigger comprehensive 4-layer checking (462ms total).

\subsection{Graceful Degradation}

When verification layers detect problems, the system responds through a graduated protocol: (1) minor structural issues trigger reasoning regeneration with a maximum of three retry attempts; (2) moderate inconsistencies activate conservative behavior modes with reduced speed and expanded safety margins; (3) physical constraint violations result in immediate safe-state transitions following ISO 10218 \cite{iso10218} emergency stop procedures; (4) repeated failures across multiple reasoning cycles escalate to human takeover. This ensures the robot remains useful even when full autonomous reasoning capability is compromised.

\section{Limitations and Future Work}

\textbf{Evaluation Scope:} Testing focused on 500 warehouse scenarios across 1,000 simulated operation hours. While the scenario mix includes adversarial cases (15\%), generalization to fundamentally different domains (outdoor delivery, aerial inspection, underwater exploration) requires dedicated evaluation campaigns as discussed in Section VI-A.

\textbf{Adversarial Robustness:} The current adversarial test set (75 scenarios) probes constraint boundaries and semantic contradictions but does not systematically evaluate against adversarial attacks on the LLM itself (prompt injection, jailbreaking). Systematic adversarial evaluation remains critical future work.

\textbf{Constraint Authoring:} The constraint specification language requires manual authoring by domain experts. Automated constraint extraction from CAD models, safety datasheets, or regulatory documents could significantly reduce deployment effort and is an active research direction.

\textbf{Computational Cost:} Self-consistency with $k$=5 samples increases inference energy costs fivefold per decision cycle. While parallel execution mitigates wall-clock latency, energy cost remains a concern for battery-powered robots. Future work will explore adaptive sampling where $k$ is dynamically adjusted based on task risk level and available computational budget.

\section{Recommendations}

\textbf{1) Standardized Safety Benchmarks:} The community should develop benchmarks for evaluating CoT reasoning safety in robotic contexts, analogous to established NLP benchmarks but incorporating physical constraint validation and temporal consistency metrics.

\textbf{2) Faithfulness as Priority:} Chen et al.'s finding \cite{chen2025reasoning} that faithfulness degrades on harder tasks is particularly concerning for complex robotic scenarios where monitoring is most needed.

\textbf{3) Self-Consistency Adoption:} Self-consistency methods \cite{wang2023selfconsistency} should become standard for robotic CoT systems, as they provide safety benefits independent of individual trace faithfulness.

\textbf{4) Defense-in-Depth:} Independent physical constraint enforcement must complement LLM-based reasoning---no single verification technique is sufficient for safety-critical applications.

\textbf{5) Updated Regulatory Frameworks:} Policymakers should develop standards addressing LLM-integrated autonomous systems, extending existing frameworks \cite{iso10218, iso15066} to cover reasoning verification requirements.

\textbf{6) Interdisciplinary Collaboration:} The NLP, robotics, and safety engineering communities must collaborate to develop comprehensive safety frameworks that bridge the gap between language model capabilities and physical safety requirements.

\textbf{Data Availability:} Implementation details, constraint specifications, and evaluation scripts will be released at the project repository upon acceptance.

\section{Conclusion}

Chain-of-Thought reasoning represents a significant opportunity for creating autonomous robots capable of sophisticated, explainable decision-making. However, realizing this potential safely requires addressing fundamental challenges related to reasoning faithfulness, physical grounding, and uncertainty quantification.

The CT-SAFR framework presented in this paper provides a structured approach for achieving trustworthy CoT reasoning, demonstrating 94.2\% hallucination detection ($n = 500$, 95\% CI: 91.8--95.9\%) with 96.4\% combined unsafe reasoning detection and 87\% reduction in safety incidents ($p < 0.001$) under controlled warehouse scenarios. Through comprehensive ablation studies, this work demonstrates that each verification layer contributes unique safety value, with the complete system achieving performance beyond what any individual mechanism provides.

This work emphasizes that safety in LLM-integrated robotics cannot be achieved through any single technique but requires defense-in-depth through complementary verification mechanisms, robust fallback behaviors, and appropriate human oversight.

\bibliographystyle{IEEEtran}

\end{document}